\documentclass[lettersize,journal]{IEEEtran}
\usepackage{amsmath,amsfonts}
\usepackage{algorithmic}
\usepackage{algorithm}
\usepackage{array}
\usepackage[caption=false,font=normalsize,labelfont=sf,textfont=sf]{subfig}
\usepackage{textcomp}
\usepackage{stfloats}
\usepackage{url}
\usepackage{verbatim}
\usepackage{graphicx}
\usepackage{cite}
\usepackage{amsmath}
\usepackage{booktabs}
\usepackage{amssymb}
\usepackage{tabularx}
\usepackage{multirow}
\usepackage[]{hyperref}
\usepackage{makecell}

\begin{document}

\title{SNNSIR: A Simple Spiking Neural Network for Stereo Image Restoration}

\author{Ronghua Xu, Jin Xie, Jing Nie, Jiale Cao, Yanwei Pang
\thanks{R. Xu and J. Xie are with the School of Big Data and Software Engineering, Chongqing University, Chongqing 400044; and are also with Shanghai Artificial Intelligence Laboratory, Shanghai 200232, China.}
\thanks{
J. Nie is with the School of Microelectronics and Communication Engineering, Chongqing University, Chongqing 400044.}
\thanks{J. Cao and Y. Pang are with the School of Electrical and Information Engineering, Tianjin University, Tianjin 300072, China.}
}

\maketitle

\begin{abstract}
Spiking Neural Networks (SNNs), characterized by discrete binary activations, offer high computational efficiency and low energy consumption, making them well-suited for computation-intensive tasks such as stereo image restoration. In this work, we propose SNNSIR, a simple yet effective Spiking Neural Network for Stereo Image Restoration, specifically designed under the spike-driven paradigm where neurons transmit information through sparse, event-based binary spikes. In contrast to existing hybrid SNN-ANN models that still rely on operations such as floating-point matrix division or exponentiation, which are incompatible with the binary and event-driven nature of SNNs, our proposed SNNSIR adopts a fully spike-driven architecture to achieve low-power and hardware-friendly computation. To address the expressiveness limitations of binary spiking neurons, we first introduce a lightweight Spike Residual Basic Block (SRBB) to enhance information flow via spike-compatible residual learning. Building on this, the Spike Stereo Convolutional Modulation (SSCM) module introduces simplified nonlinearity through element-wise multiplication and highlights noise-sensitive regions via cross-view-aware modulation. Complementing this, the Spike Stereo Cross-Attention (SSCA) module further improves stereo correspondence by enabling efficient bidirectional feature interaction across views within a spike-compatible framework. Extensive experiments on diverse stereo image restoration tasks, including rain streak removal, raindrop removal, low-light enhancement, and super-resolution demonstrate that our model achieves competitive restoration performance while significantly reducing computational overhead. These results highlight the potential for real-time, low-power stereo vision applications. The code will be available after the article is accepted.
\end{abstract}


\section{Introduction}
\IEEEPARstart{I}{mage} restoration aims to reconstruct high-quality images from degraded or low-quality inputs, including those affected by adverse weather (e.g., rain), insufficient lighting (e.g., low-light environments), or limited resolution. Stereo image restoration extends this task by leveraging a pair of degraded left and right images to produce more detailed and geometrically consistent results. One of its key advantages lies in the interactive information exchange between the two views, which allows the model to recover scene details that are degraded or missing in one image using cues from its counterpart. By exploiting such complementary information, stereo restoration often surpasses single-image approaches in both visual fidelity and structural coherence.

Additionally, this capability is especially critical in applications such as stereo matching~\cite{chen2022pseudo,wu2023semi}, depth estimation~\cite{brucker2024cross}, and 3D perception tasks like object tracking~\cite{li2020joint} and detection~\cite{wu2023semi,chen2022pseudo}. Moreover, high-quality stereo image restoration serves as a fundamental component in robot vision and embodied intelligence systems, where accurate spatial perception and robust visual input are essential for decision making, navigation, and interaction with dynamic environments.

\begin{figure}[t!]
\centering
\includegraphics[width=\linewidth]{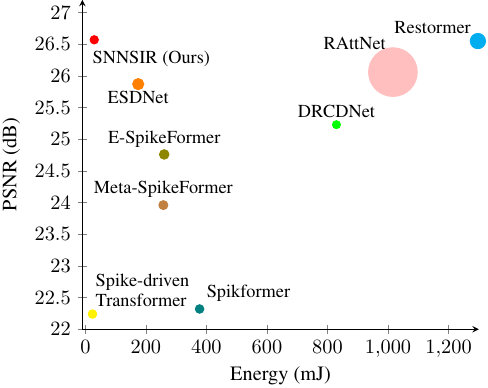}
\caption{Comparison of PSNR (dB) vs. Energy Consumption (mJ) on the SeteroWaterdrop test set. Circle size denotes model size. Our SNNSIR achieves competitive restoration performance with significantly lower energy consumption and model size.}
\label{intro_fig}
\end{figure}

Cross-view interaction lies at the core of stereo restoration tasks. Recent stereo restoration methods~\cite{wang2019learning,yan2020disparity,wang2021symmetric,chu2022nafssr,zhang2022beyond,nie2023context,wang2025apanet} have commonly enhanced such interaction through guided alignment, parallax attention, and semantic fusion, achieving notable improvements in restoration quality. However, these approaches typically rely on dual-branch encoders, and computationally intensive interaction modules, leading to considerable computational overhead. This complexity hinders real-time deployment on resource-limited platforms such as robots, UAVs~\cite{chang2024uav,feng2024hazydet}, and edge devices. Consequently, there is an increasing demand for lightweight stereo restoration frameworks that preserve the benefits of cross-view modeling while ensuring computational efficiency.

To address the growing need for computational efficiency in vision tasks, recent research has explored Spiking Neural Networks (SNNs)~\cite{maass1997networks}, often referred to as the third generation of neural networks. Unlike ANNs, SNNs operate using discrete spikes triggered by threshold-based membrane potentials, offering sparse, event-driven computation and inherent temporal modeling. These characteristics enable significant reductions in energy consumption and computational cost.
SNNs have shown promising results in high-level tasks such as image classification~\cite{fang2021deep,zhou2022spikformer,yao2023spike,shi2024spikingresformer,yao2024spike}, object detection~\cite{luo2024integer}, and semantic segmentation~\cite{yao2025scaling}. Notably, ESDNet~\cite{song2024learning} recently demonstrated the potential of SNNs in rain removal, achieving effective restoration with low energy consumption.
These developments point to the potential of SNNs as a lightweight solution for stereo image restoration, particularly on resource-constrained platforms. Despite their strengths in temporal modeling and computational efficiency, SNNs have yet to be fully explored in this domain. 
Moreover, existing SNN-based single-image restoration methods, such as ESDNet~\cite{song2024learning}, adopt hybrid SNN-ANN architectures that rely on non-spike-compatible operations, including floating-point division and sigmoid activations, thereby limiting their energy efficiency. For example, the Mixed Attention Unit (MAU) in ESDNet employs sigmoid functions, underscoring the lack of exploration into achieving nonlinear representation within fully spike-driven frameworks.

\begin{figure}[t]
	\centering
	\includegraphics[width=\linewidth]{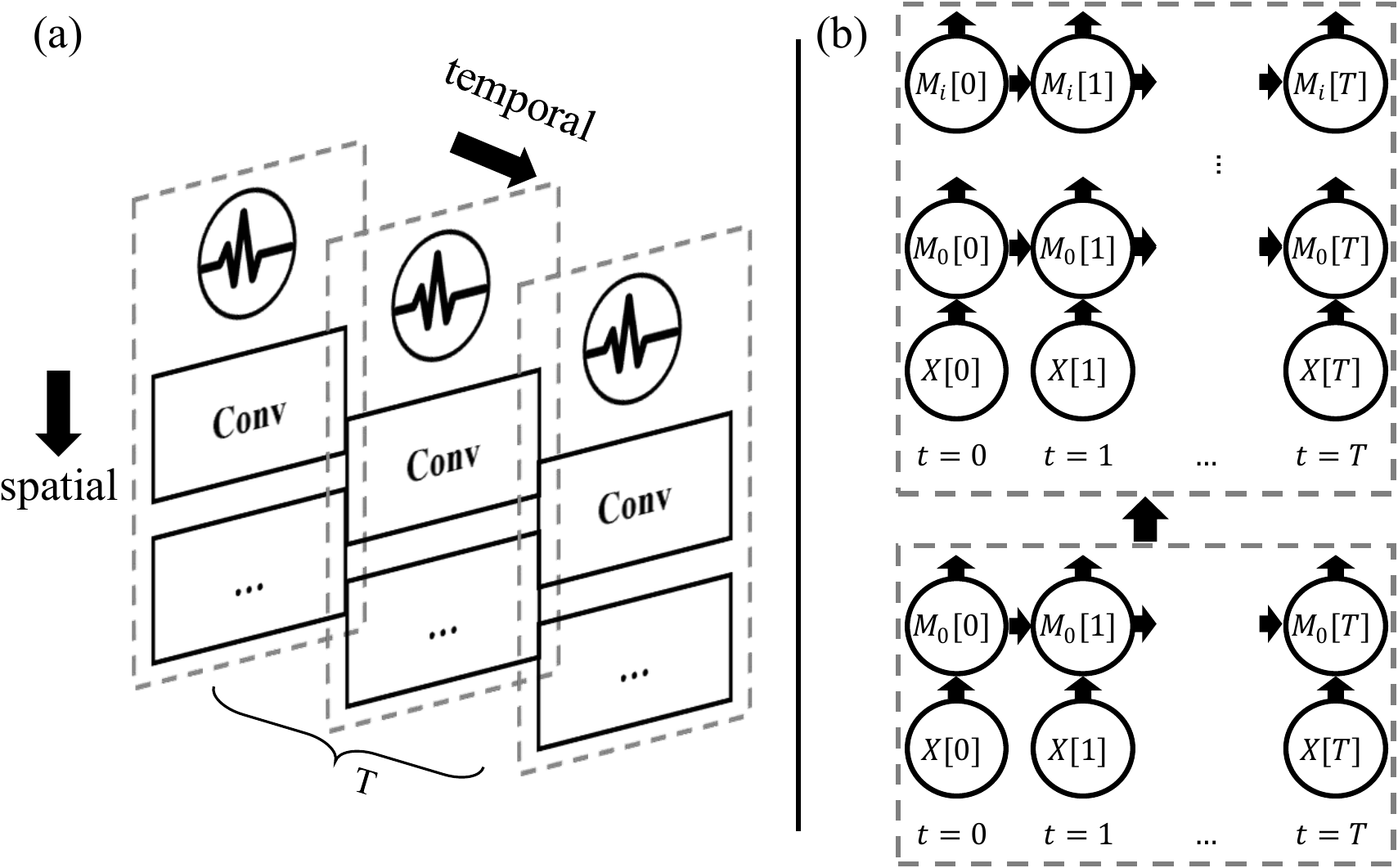}
	\caption{(a) and (b) respectively illustrate the spatiotemporal characteristics of SNN in terms of network architecture and data propagation.}
	\label{fig:spatiotemporal}
\end{figure}

In this paper, we propose a simple spiking neural network for stereo image restoration, called SNNSIR. The architecture adopts a coarse-to-fine framework, enabling hierarchical refinement while maintaining a fully spike-driven processing pipeline. Unlike ANNs that depend on floating-point operations and nonlinear activations, SNNSIR achieves nonlinear representation and cross-view reasoning using spike-compatible modules. 

To realize this design, we introduce several SNN-based modules. To enhance the representational capacity of binary spike neurons, we deisgn the Spike Residual Basic Block (SRBB), which integrates a residual learning scheme tailored for SNNs. Two SRBBs form a Feature Extraction Block (FEB) that benefits from  the discrete computation nature to avoid excessive computational burden. 
To introduce nonlinearity into the SNN model, we draw inspiration from prior findings~\cite{chen2022simple} showing that element-wise multiplication can serve as an effective substitute for traditional activations. Based on this, we design the Spike Stereo Convolutional Modulation (SSCM) module, which employs multiplication as a spike-compatible nonlinear activation. Additionally, its channel- and spatial-wise modulation enhances feature representation and highlights degraded regions to facilitate subsequent restoration.
To capture long-range dependencies and enhance cross-view interaction, we introduce the Spike Stereo Cross-Attention (SSCA) module, which facilitates efficient inter-view information exchange within a spike-driven framework. 
In addition, to compensate for the loss of fine-grained details in deeper layers, a coarse-to-fine architecture is adopted, where lightweight Spike Stereo Refinement Blocks (SSRBs) are introduced in the final stage for local restoration.
Furthermore, to enable direct training of the SNN-based model, we adopt surrogate gradient techniques~\cite{kim2020unifying}, which effectively address the non-differentiability of spike functions. Our model achieves comparable restoration performance to existing ANN- and SNN-based methods, while significantly reducing energy consumption.

The main contributions of this paper are as follows:
\begin{itemize}
\item We propose a simple spiking neural network for stereo image restoration, termed SNNSIR, offering a novel and energy-efficient perspective for adapting stereo restoration to real-world, resource-constrained devices.

\item We design the Spike Stereo Cross-Attention (SSCA) module, Spike Stereo Convolutional Modulation (SSCM) module, and Spike Stereo Refinement Block (SSRB). SSCM introduces nonlinearity and guides the network to focus on degraded regions.
SSCA enables efficient long-range cross-view interaction.
SSRB refines local details in a lightweight manner.

\item Extensive experiments on stereo image raindrop removal, rain streak removal, low-light enhancement, and super-resolution validate the effectiveness of our method. SNNSIR achieves comparable performance to representative ANN-based methods while reducing energy consumption by up to 97.73\%, and consistently outperforms existing SNN-based approaches. To our knowledge, this work establishes the first dedicated and high-performing baseline for stereo image restoration using spiking neural networks.
\end{itemize}

\begin{figure*}[ht]
  \centering
  \includegraphics[width=\linewidth]{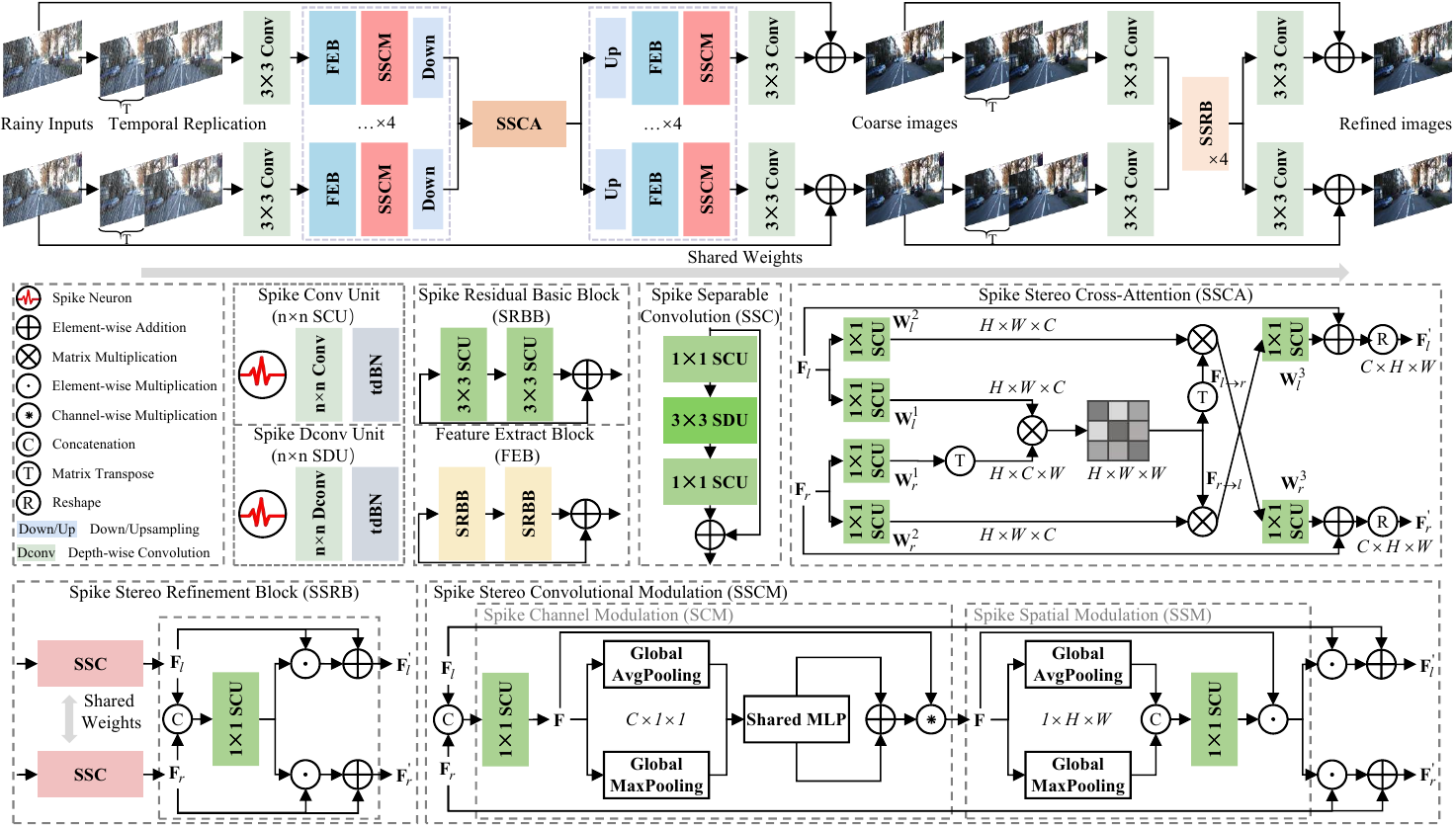}
\caption{
The architecture of the proposed two-stage spike-driven stereo image restoration framework. In the first stage, a coarse restored image is generated, followed by a lightweight refinement stage that removes artifacts to produce the final output. The framework incorporates several core components: a Feature Extraction Block (FEB) with two Spike Residual Basic Blocks (SRBBs) for efficient feature extraction, a Spike Stereo Convolutional Modulation (SSCM) module to enhance nonlinearity, and a Spike Stereo Cross Attention (SSCA) module  to facilitate cross-view interaction.
}
  \label{fig:architecture}
\end{figure*}

\section{Preliminary}
\label{sec:preliminary}

\textbf{Spatiotemporal Nature.} SNNs are spatiotemporal. In Figure~\ref{fig:spatiotemporal}(a), we follow~\cite{hu2024advancing}, where the spatial dimension is represented by the vertical stacking of convolutional layers (Conv) within each time step, capturing spatial dependencies in the input, and the temporal dimension is depicted by the sequential propagation of neural activities across different time steps $T$, emphasizing the dynamic nature of SNNs. Specifically, the feature propagates layer by layer along the temporal dimension which is represented in Figure~\ref{fig:spatiotemporal}(b), as described by~\cite{Fang2023SpikingJelly}, where $\mathbf{X}$ and $\mathbf{M}$ denote input features and intermediate result in each layer and time step with total $T$, separately Therefore, for static images, we first temporally replicate them over $T$ time steps to generate a sequence $\mathbf{X}[T]$, as shown in Figure~\ref{fig:architecture}.
Considering the biological characteristics of Leaky Integrate-and-Fire (LIF) neuron~\cite{maass1997networks} that can transform input features into binary (0/1) spike sequences and its ease of simulation on computers, we select it to implement the propagation of spike signals in the network. Its dynamic equation can be described as:

\begin{equation}
  \mathbf{U}_{i}[t]=\mathbf{V}_{i}[t-1]+\frac{1}{\tau}(\mathbf{X}_{i}[t]-(\mathbf{V}_{i}[t-1]-u_{\text {rest}})),
  \label{eq:charge}
\end{equation}
\begin{equation}
\mathbf{S}_i[t] = {\theta}(\mathbf{U}_i[t] - u_{\text{th}}),
\label{eq:firing}
\end{equation}
\begin{equation}
\theta(x) = \begin{cases}1, & x\geq0 \\0, & x < 0\end{cases},
\label{eq:stepFunction}
\end{equation}
\begin{equation}
\mathbf{V}_i[t] = (1 - \mathbf{S}_i[t])\mathbf{U}_i[t] + \mathbf{S}_i[t]u_{\text{rest}},
\label{eq:reset}
\end{equation}
where $t$ and $i$ represents the $t$-th time step and the $i$-th spike neuron. The whole procedure encompasses three distinct phases: charging, firing, and resetting. Equation~\eqref{eq:charge} represents the charging progression, where the interaction between spatial input $\mathbf{X}_{i}[t]$ and the membrane potential $\mathbf{V}_{i}[t-1]$ at the last moment generates the current membrane potential $\mathbf{U}_{i}[t]$. ${\tau}$ is the membrane time constant. Equation~\eqref{eq:firing} represents the firing progression. If the membrane potential after charging exceeds the threshold potential $u_{\text{th}}$, a spike is emitted through the step function in Equation~\eqref{eq:stepFunction}. Otherwise, the output spike is 0. After firing, reset the membrane potential which is described in Equation~\eqref{eq:reset}. Here, $u_{\text{rest}}$ and $\mathbf{V}_i[t]$ represent the reset potential and the final potential of this neuron.

\textbf{Energy Consumption.} In ANNs, floating-point operations (FLOPs) are commonly used to evaluate the computational cost, where most of the FLOPs are multiply-accumulate operations (MACs). However, in SNNs, which handle binary matrices, the operations reduce to only accumulation operations (ACs), referred to as synaptic operations (SOPs)~\cite{zhou2022spikformer,yao2023spike}. We assume that the implementation of MAC and AC operations is carried out on 45nm hardware~\cite{horowitz20141}, with the energy consumption $E_{\text{MAC}} = 4.6pJ$ and $E_{\text{AC}} = 0.9pJ$.
For an input $\mathbf{X}$, We follow \cite{zhou2022spikformer}, defining SOPs as follows: 
\begin{equation}
	\text{SOPs}(\mathbf{X}) = T \times fr \times \text{FLOPs}(\mathbf{X}),
\end{equation}
where $T$ is the time step and $fr$ denotes the firing rate. Here, the product of $fr$ and $\text{FLOPs}(\mathbf{X})$ corresponds to the number of 1s in the feature matrix after processing by the spike neurons.

For our proposed network, to be more precise, we describe its energy consumption as follows:
\begin{equation}
	E = 0.9pJ\times \sum\text{SOPs}(\mathbf{X}_s) + 4.6pJ \times \sum\text{FLOPs}(\mathbf{X}_a)
\end{equation}
Where $\mathbf{X}_s$
is the binary spike matrix processed by the spike neurons. $\mathbf{X}_a$ is the floating-point matrix. 

At the code level, calculating SOPs is equivalent to counting the number of 1s in the output from the spike neurons shown in Figure~\ref{fig:spatiotemporal}(a), which corresponds to the number of spikes emitted. Since the spike firing rate varies for each dataset, we first calculate the total number of spikes in the test set for each dataset, and then compute the ratio of the total spikes to the number of samples. This provides us the SOPs per sample for that dataset, from which the energy consumption can be derived. As we all know, FLOPs are fixed and only need to account for the computation that occurs in non-spike sequences. Taking our proposed SNNSIR as an example, the only non-spike sequences are the $3\times3$ convolutional layers at the beginning and end of the degradation removal and refinement stages, while the rest of the operations are spike sequences. In this paper, SOPs and FLOPs are measured on $256\times256$, except for super-resolution, which uses a size of $64\times64$. The results of FLOPs and SOPs for different tasks on different datasets are shown in Table~\ref{tab:compution}. For tasks involving multiple datasets, the computational complexity is taken as the average across all datasets for that task.

\section{Method}

The overall architecture of the proposed SNNSIR is illustrated in Figure~\ref{fig:architecture}. SNNSIR restores a degraded stereo image pair through a coarse-to-fine framework composed of a U-shaped encoder-decoder and a lightweight refinement stage.

The first stage adopts a 5-layer U-shaped encoder-decoder with channel dimensions [32, 64, 96, 128, 160], enabling multi-scale feature representation with a simple structure. Given a static stereo input of size $3 \times H \times W$, temporal replication produces inputs $\mathbf{X}_l,\mathbf{X}_r \in \mathbb{R}^{T \times 3 \times H \times W}$ for the left and right views. These replicated inputs are then passed through a $3 \times 3$ convolution to extract shallow features $\mathbf{F}_{l},\mathbf{F}_{r} \in \mathbb{R}^{T \times C \times H \times W}$, serving as the initial representation for subsequent processing.

These features are processed by the Feature Extraction Block (FEB) for single-view feature extraction, followed by the Spike Stereo Convolutional Modulation (SSCM) module, which introduces nonlinearity and highlight noise-sensitive region. Feature maps are progressively downsampled, reducing spatial resolution while increasing channel depth. The encoded features are then fed into the Spike Stereo Cross-Attention (SSCA) module to model cross-view long-range dependencies. The decoder mirrors the encoder and includes skip connections via element-wise addition.
After decoding, a $3 \times 3$ convolution and temporal average pooling generate the left and right residual maps$\mathbf{O}_{l},\mathbf{O}_{r} \in \mathbb{R}^{3 \times H \times W}$, which are added to the original inputs to obtain coarse predictions. 

Finally, to recover fine details, a refinement stage composed of four Spike Stereo Refinement Blocks (SSRBs) operates at a fixed resolution with 32 channels, avoiding information loss from further downsampling.

Next, we provide detailed descriptions of the core modules in SNNSIR.

\noindent\textbf{Spike Convolution Unit (SCU) and Spike Depth-wise Convolution Unit(SDU).} 
Given that SNNs process binary spike sequences, their foundational  structure~\cite{fang2021deep,zheng2021going,zhou2022spikformer,luo2024integer,shi2024spikingresformer,song2024learning,yao2024spike} typically consists of a LIF neuron~\cite{maass1997networks} followed by a convolutional or linear layer. tdBN~\cite{zheng2021going} performs normalization simultaneously in both spatial and temporal dimensions that can alleviate the gradient and firing rate issues in deep SNN training. ~\cite{song2024learning} proposed a Spike Convolution Unit (SCU), but it is not specific enough. We design $n \times n$ SCU and $n \times n$ Spike Depth-wise Convolution Unit ($n \times n$ SDU) based on the kernel size and convolution type. These two structures are described as:
\begin{equation}
	\begin{aligned}
		\mathbf{X}_{i+1}^{SCU}[t] &= \text{tdBN} ( \text{Conv} ( \mathbf{S}_i[t]) ),\\
		\mathbf{X}_{i+1}^{SDU}[t] &= \text{tdBN} ( \text{Dconv} ( \mathbf{S}_i[t]) ),
	\end{aligned}
\end{equation}
where the definition of $\mathbf{S}$, $i$ and $t$ can be found in Section~\ref{sec:preliminary}.

\noindent\textbf{Spike Residual Basic Block (SRBB).} To ensure sufficient network depth while preserving the ability to extract rich local features, we adopt the residual basic block ~\cite{he2016identity} as the fundamental building unit of our network. Unlike vanilla residual basic block, we replace the standard convolutional layers with $3 \times 3$ SCUs to better accommodate the characteristics of SNN. Compared to traditional ANNs, implementing residual connections in SNNs requires specialized designs. Among existing strategies~\cite{fang2021deep,zheng2021going,hu2024advancing}, we adopt the Membrane Shortcut (MS)~\cite{hu2024advancing} as it best preserves identity mapping within the spike-driven paradigm, forming our Spike Residual Basic Block (SRBB). We construct each FEB by stacking two SRBBs to learn richer, hierarchical spatiotemporal representations. FEBs' weights are shared across the left and right stereo views to reduce the parameter count and enforce consistent feature extraction.


\noindent\textbf{Spike Stereo Convolutional Modulation (SSCM) module.} To address the limited non-linearity and lack of cross-view interaction in spiking neural networks, the proposed Spike Stereo Convolutional Modulation (SSCM) module introduces spike-compatible non-linear activation while enabling stereo-aware feature refinement through convolutional modulation. Inspired by~\cite{chen2022simple}, SSCM employs element-wise multiplication to introduce lightweight non-linearity without relying on conventional activation functions such as sigmoid, which are incompatible with the spike-driven paradigm.

As illustrated in Figure~\ref{fig:architecture}, SSCM takes the concatenated left and right view features as input to form a fused representation $\mathbf{F}$, which encodes coarse inter-view correlations and enables the network to leverage global context for compensating view-specific degradations caused by noise or occlusion. It then sequentially applies Spike Channel Modulation (SCM) and Spike Spatial Modulation (SSM), which modulate feature responses along the channel and spatial dimensions, respectively. Both submodules rely on element-wise multiplication to introduce non-linearity. The detailed computational process is defined as:

\begin{equation}
	\begin{aligned}
		\text{SCM}(\mathbf{F})&=\mathbf{F}\circledast(\mathbf{W}(\text{GAP}(\mathbf{F})) + \mathbf{W}(\text{GMP}(\mathbf{F}))),\\
		\text{SSM}(\mathbf{F})&=\mathbf{F}\odot(\text{SCU}([\text{GAP}(\mathbf{F}),\text{GMP}(\mathbf{F})])),\\
		\mathbf{F}^{'}_{l/r}&=\text{SSM}(\mathbf{F})\odot\mathbf{F}_{l/r} + \mathbf{F}_{l/r},\\
	\end{aligned}
	\label{eq:SSCBAM}
\end{equation}
where $\text{GAP}(\cdot)$ and $\text{GMP}(\cdot)$ denote global average pooling and global max pooling, respectively; $\mathbf{W}(\cdot)$ is a shared linear operation; $\circledast$ and $\odot$ represent channel- and spatial-wise element-wise multiplication, respectively.
The final outputs, $\textbf{F}^{'}_l$ and $\textbf{F}^{'}_r$, are obtained through residual modulation based on spike-compatible operations.
In summary, SSCM acts as a unified component that simultaneously enhances non-linearity and facilitates cross-view information integration, playing a key role in high-quality stereo restoration under the constraints of biologically inspired spiking computation.

\noindent\textbf{Spike Stereo Cross-Attention (SSCA) module.} The Spike Stereo Cross-Attention (SSCA) module, a SNN variant of the Stereo Cross-Attention Module~\cite{chu2022nafssr}, is designed to enhancing feature interaction between the left and right images provided by stereo vision systems. Its key modification lies in the introduction of the SCU and the removal of the activation function, making it spike-compatible. Due to the characteristics of stereo vision, horizontal disparities are present while there is no significant disparity in the vertical direction. As a result, performing attention computation in the horizontal direction is more efficient. 
\begin{equation}
	\begin{aligned}
		\mathbf{F}^{'}_l &= \mathbf{W}_l^3 (\mathbf{W}_l^1 \mathbf{F}_l \times (\mathbf{W}_r^1 \mathbf{F}_r)^T \times \mathbf{W}_r^2 \mathbf{F}_r) + \mathbf{F}_l, \\
		\mathbf{F}^{'}_r &= \mathbf{W}_r^3 ((\mathbf{W}_l^1 \mathbf{F}_l \times (\mathbf{W}_r^1 \mathbf{F}_r)^T)^T \times \mathbf{W}_l^2 \mathbf{F}_l) + \mathbf{F}_r.
	\end{aligned}
	\label{eq:SSCAM}
\end{equation}
In Equation~(\ref{eq:SSCAM}), feature maps $\mathbf{F}_l$ and $\mathbf{F}_r$ are first reshaped into $H\times W\times C$ dimensions. The reshaped features are then processed by the learned weight matrices, 
$\mathbf{W}_l^1$, 
$\mathbf{W}_l^2$, 
$\mathbf{W}_r^1$, 
$\mathbf{W}_r^2$, $\mathbf{W}_l^3$ and $\mathbf{W}_r^3$ from $1\times1$ SCU. The output features 
$\mathbf{F}^{'}_l$ and $\mathbf{F}^{'}_r$ are refined by combining the spatial information from both views using the cross-attention computation and reshaped back to the 
$C \times H\times W$ dimensions.

\begin{table*}[t!]
	\caption{Comparison for stereo image raindrop removal on the Stereo Waterdrop dataset. P and E indicate the number of parameters and energy consumption (mJ), respectively.}
	\label{tab:quantitiveStereo}
	\setlength{\tabcolsep}{1mm}
	\centering
	{
		\fontsize{9}{10}\selectfont
		\begin{tabular}{cccccccc}
			\toprule
			Methods&Type&PSNR~(dB) $\uparrow$&MS-SSIM $\uparrow$&LPIPS $\downarrow$ &P~(M) $\downarrow$&E~(mJ) $\downarrow$&Times~(s) $\downarrow$\\
			\midrule
			Pix2Pix&\multirow{7}{*}{ANN}&  22.76 & 0.895 & 0.217 & 54.41 &-&\textbf{0.045}\\
			Qian et al.&  & 24.47& 0.900&0.163& 6.24 &822.39&0.083\\
			Liu et al.&  &22.70&0.833&0.247&16.69&-&1.241\\
			Quan et al.&  &24.97&0.913&0.153&7.27&-&0.115\\
			DRCDNet&  &25.23& 0.906 & 0.176 & \textbf{2.25} & 829.64&0.173 \\
			Restormer& & \textbf{26.55} &0.945 & 0.124 & 26.13 &1297.11&0.439 \\
			RAttNet&&26.06&\textbf{0.950}&0.096&136.55&1016.30&0.210\\
			\hline
			Spikformer&\multirow{6}{*}{SNN}&22.32&0.828&0.186 &4.04&377.45 &0.680 \\
			Spike-driven Transformer&&22.24&0.830& 0.182&4.04 &\textbf{23.53} & 0.642\\
			Meta-SpikeFormer&&23.96&0.874&0.151 &5.54& 257.74& 0.663\\
			E-SpikeFormer&&24.76&0.918& 0.102&7.30&260.61 & \textbf{0.032}\\
			ESDNet&&25.87&0.941&\textbf{0.072}&12.81&174.63 &16.148\\
			SNNSIR(Ours)&&\textbf{26.57}&\textbf{0.949}&\textbf{0.062}&\textbf{3.26}&\textbf{29.32}&0.411\\
			\bottomrule
		\end{tabular}
	}
\end{table*}

\begin{figure*}[t!]
	\centering
	\includegraphics[width=\linewidth]{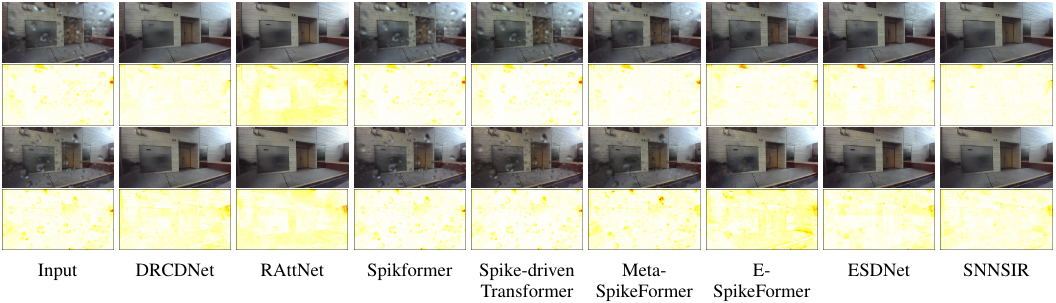}
	\caption{Visual comparison for the stereo image raindrop removal task on the Stereowaterdrop dataset. The top and bottom rows show the left and right views, respectively. Best viewed by zooming in for details.}
	\label{fig:visualWaterdrop}
\end{figure*}

\noindent\textbf{Spike Stereo Refinement Block (SSRB).} 
To compensate for the loss of fine details in the coarse output, we introduce a refinement stage that operates at full resolution to enhance textures and residual information without further downsampling. In Figure\ref{fig:architecture}, we design the SSRB, which consists of a Spike Separable Convolution (SSC) inspired by~\cite{yao2024spike}, adapted from the inverted separable convolution~\cite{sandler2018mobilenetv2} for efficient spatial feature extraction. As shown in the Figure~\ref{fig:architecture}, after being handled by the SSC, the features from both views are first concatenated and processed by a SCU, whose output is then used to modulate each branch via element-wise multiplication and Membrane Shortcut~\cite{hu2024advancing}, enhancing discriminative feature refinement in a lightweight manner while providing certain inter-view interaction and non-linearity.

\subsection{Loss Functions}
\label{sec:loss}
Due to the distinct tasks undertaken by the two stages of the network, different loss functions are employed for each stage. The $L_1$ loss emphasizes pixel-level accuracy, making it suitable for the first stage focused on degradation removal. Perception loss~\cite{johnson2016perceptual,chen2017photographic} , which calculates the loss using features extracted from multiple layers to align both local high-frequency details and global low-frequency structures, is employed in the refinement stage and denoted as $L_p$. $L_1$ and $L_p$ are computed as:
\begin{equation}
	\begin{aligned} 
		L_1&=\frac{1}{N}\sum_{i=1}^{N}|\mathbf{X}_{l,i}-\mathbf{O}_{l,i}| + \frac{1}{N}\sum_{i=1}^{N} |\mathbf{X}_{r,i}-\mathbf{O}_{r,i}|,\\
		L_p&=\sum_k \lambda_k(\|\mathbf{X}_l^k-\mathbf{O}_r^k\|_1+\|\mathbf{X}_r^k-\mathbf{O}_r^k\|_1),
	\end{aligned}
\end{equation}
where $\mathbf{X}$ and $\mathbf{O}$ are input and output, $\mathbf{X}_{l,i}$ means the $i$-th pixel with a total number of N in the left view and similarly for the right view, $k$ denotes the selected layers, $\| \cdot \|_1$ denotes the $l_1$ norm, and $\lambda_k$ functions as a hyperparameter to adjust the contribution of each layer. Thus, the overall loss is the sum of the two losses, expressed as $L_{total} = L_1 + L_p$.

\section{Experiments}
We conducted extensive analyses on three common tasks: \textbf{(a)} stereo image deraining, \textbf{(b)} stereo image low-light enhancement, and \textbf{(c)}  stereo image super-resolution. Details of the datasets, training hyperparameters for each task, and more experimental results are provided in the supplementary material. In the result tables of this section, the best and second-best results are both highlighted in bold.

\subsection{Implementation Details}
\label{sec:ImpDetails}
Our proposed network adopts a coarse-to-fine architecture, where the degradation removal stage uses channel dimensions of $[32, 64, 96, 128, 160]$ and in the refinement stage, four SSRBs are employed with an embedding dimension of 32. AdamW optimizer~(${\beta}_1=0.9$, ${\beta}_2=0.99$, $\text{weight~decay}=1e-4$) is used with a fixed learning rate $1e-3$. Furthermore, since low-light enhancement degradation is not high-frequency in nature, we set its spike activation threshold to 0.1, while 0.2 for other tasks. The default number of time steps $T$ for all tasks is 4. Due to the significant differences across datasets for each task, other hyperparameters vary considerably. Detailed configurations are provided in the supplementary material. Our network is implemented via SpikingJelly~\cite{Fang2023SpikingJelly}, a PyTorch-based SNN framework. All experiments are carried out on an NVIDIA GeForce RTX 3090 (24GB, 350W).

\noindent\textbf{Evaluation Metrics and Visualization.} To evaluate restoration performance, we adopt PSNR, SSIM, MS-SSIM~\cite{wang2003multiscale}, and LPIPS~\cite{zhang2018unreasonable} as evaluation metrics. The number of parameters~(P), computational complexity, and energy consumption~(E) are used to evaluate the computational cost and resource overhead. Specifically, the computational complexity includes FLOPs and SOPs. The introduction to these two metrics can be found in Section~\ref{sec:preliminary}. In this paper,  FLOPs and SOPs are measured on an image size $256\times256$, except for the super-resolution task, where a size of $64\times64$ is used. For visual comparison, we use error maps that intuitively depict the pixel-wise differences between the restored images and the ground-truth images through a red-yellow-white color gradient. In these error maps, larger errors correspond to more intense red hues, smaller errors appear closer to white, and moderate errors are represented by yellow. 

\noindent\textbf{Comparison Methods.} For the stereo raindrop removal task, ANN-based methods including DRCDNet \cite{wang2023rcdnet}, Restormer \cite{zamir2022restormer}, and RAttNet \cite{shi2021stereo} are retrained based on official code, while results of Pix2Pix~\cite{isola2017image}, Qian et al.~\cite{qian2018attentive}, Liu et al. ~\cite{liu2020learning} and Quan et al.~\cite{quan2019deep} are referenced from \cite{nie2023context}. Given the scarcity of SNN-based restoration methods, Spikformer\cite{zhou2022spikformer}, Spike-driven Transformer\cite{yao2023spike}, Meta-SpikeFormer\cite{yao2024spike}, and E-SpikeFormer\cite{yao2025scaling} are adapted from state-of-the-art SNN-based classification methods. 
Each method follows a 5-layer U-shaped encoder-decoder structure consistent with SNNSIR. Each layer uses the core block from its original structure, with the channel numbers adjusted to match those of SNNSIR, and the remaining parameters of each block are accordingly modified. 
And ESDNet\cite{song2024learning} is a dedicated monocular restoration method. These five SNN-based methods are retrained on all tasks. Results of ANN-based methods including DID-MDN~\cite{zhang2018density}, DeHRain~\cite{li2019heavy}, ZeroDCE~\cite{guo2020zero}, RetinexNet~\cite{wei2018deep} and DRBN~\cite{yang2020fidelity} for stereo image rain streak removal and low-light enhancement are respectively referenced from \cite{nie2023context} and \cite{huang2022low}. Notably, except for RAttNet and our SNNSIR, all other methods are monocular. To be fair, monocular methods are trained separately on the left and right views. Apart from parameter sharing, all evaluation metrics are computed in the same manner as those for stereo methods. Specifically, the restoration quality is averaged across the left and right views, while the computational resource consumption is the sum of the resources required for both views.

\begin{table}[t]
	\caption{Comparison for stereo image rain streak removal on the RainKT12 and RainKT15 datasets.}
	\label{tab:quantitiveKT}
	\setlength{\tabcolsep}{1mm}
	\centering
	{
		\fontsize{9}{10}\selectfont
		\begin{tabular}{ccccc}
			\toprule
			Methods &P~(M)&E~(mJ)& RainKT12& RainKT15\\
			\midrule
			DID-MDN &\textbf{0.47}& 93.23&29.14/0.901& 28.97/0.899  \\
			DeHRain&50.33&2486.78&\textbf{31.02}/\textbf{0.923} & \textbf{30.84}/\textbf{0.921} \\
			\midrule
			Spikformer&4.04&1255.57 &18.73/0.600&22.28/0.662 \\
			Spike-driven-&\multirow{2}{*}{4.04}&\multirow{2}{*}{\textbf{41.70}} &\multirow{2}{*}{23.46/0.703} &\multirow{2}{*}{24.32/0.728} \\
			Transformer& & & &\\
			Meta-SpikeFormer&5.54&201.45 &26.78/0.846&27.12/0.847\\
			E-SpikeFormer&7.30&281.91&21.65/0.709&21.91/0.713\\
			ESDNet&12.81&217.13 &28.52 /0.887 & 28.80/0.884 \\
			SNNSIR&\textbf{3.26}& \textbf{32.69}&\textbf{30.97}/\textbf{0.922}  & \textbf{31.43}/\textbf{0.920}\\
			\bottomrule
		\end{tabular}
	}
\end{table}

\begin{table}[t!]
	\caption{Comparison for stereo image low-light enhancement on the Middlebury and Holopix50k datasets.}
	\label{tab:quantitiveLLE}
	\centering
	\setlength{\tabcolsep}{1mm}
	{
		\fontsize{9}{10}\selectfont
		\begin{tabular}{ccccc}
			\toprule
			Methods&P~(M)&E~(mJ)& Middlebury &Holopix50k \\
			\midrule
			ZeroDCE & \textbf{0.08}&98.53& 15.43/0.715& 13.28/0.652  \\
			RetinexNet & 0.84& 325.73&22.66 0.800 &  19.20/0.779\\
			DRBN & \textbf{0.58} &173.83& \textbf{31.02}/\textbf{0.943} & \textbf{25.09}/\textbf{0.903}\\
			\midrule
			Spikformer&4.04& 486.19&21.99/0.770 &17.49/0.678 \\
			Spike-driven-&\multirow{2}{*}{4.04}& \multirow{2}{*}{\textbf{48.95}}&\multirow{2}{*}{26.84/0.800 }&\multirow{2}{*}{20.54/0.787}\\
			Transformer& & & &\\
			Meta-SpikeFormer&5.54& 301.30&22.47/0.792 &19.54/0.787\\
			ESDNet&12.81& 208.55 & 30.19/\textbf{0.928}& 23.75/0.883 \\
			SNNSIR  & 3.26 & \textbf{36.48} & \textbf{31.28}/0.923&  \textbf{24.82}/\textbf{0.888}  \\
			\bottomrule
		\end{tabular}
	}
\end{table}

\begin{table}[t!]
	\caption{Comparison for stereo image super-resolution on the Middlebury and Flickr1024 datasets.}
	\label{tab:SRResults}
	\centering
	\setlength{\tabcolsep}{1mm}
	{
		\fontsize{9}{10}\selectfont
		\begin{tabular}{ccccccc}
			\toprule
			Methods&P~(M) &E~(mJ) & Middlebury & Flickr1024 \\
			\midrule
			Spikformer&\textbf{0.12}&8.27&26.59/0.747&21.91/0.614 \\
			Spike-driven-&\multirow{2}{*}{\textbf{0.12}}&\multirow{2}{*}{\textbf{1.31}}&\multirow{2}{*}{26.83/\textbf{0.757}}&\multirow{2}{*}{\textbf{22.06}/\textbf{0.627}}\\
			Transformer& & & & \\
			Meta-SpikeFormer&0.39&17.34&\textbf{26.90}/0.755&22.03/0.621\\
			E-SpikeFormer&0.45& 50.77&26.51/0.745&21.88/0.612 \\
			ESDNet&1.41& 20.92&15.36/0.663 &15.63/0.547\\
			SNNSIR & 0.33&\textbf{1.59} &\textbf{27.38}/\textbf{0.772} &\textbf{22.29}/\textbf{0.640}  \\
			\bottomrule
		\end{tabular}
	}
\end{table}

\begin{figure*}[t]
	\centering
	\includegraphics[width=\linewidth]{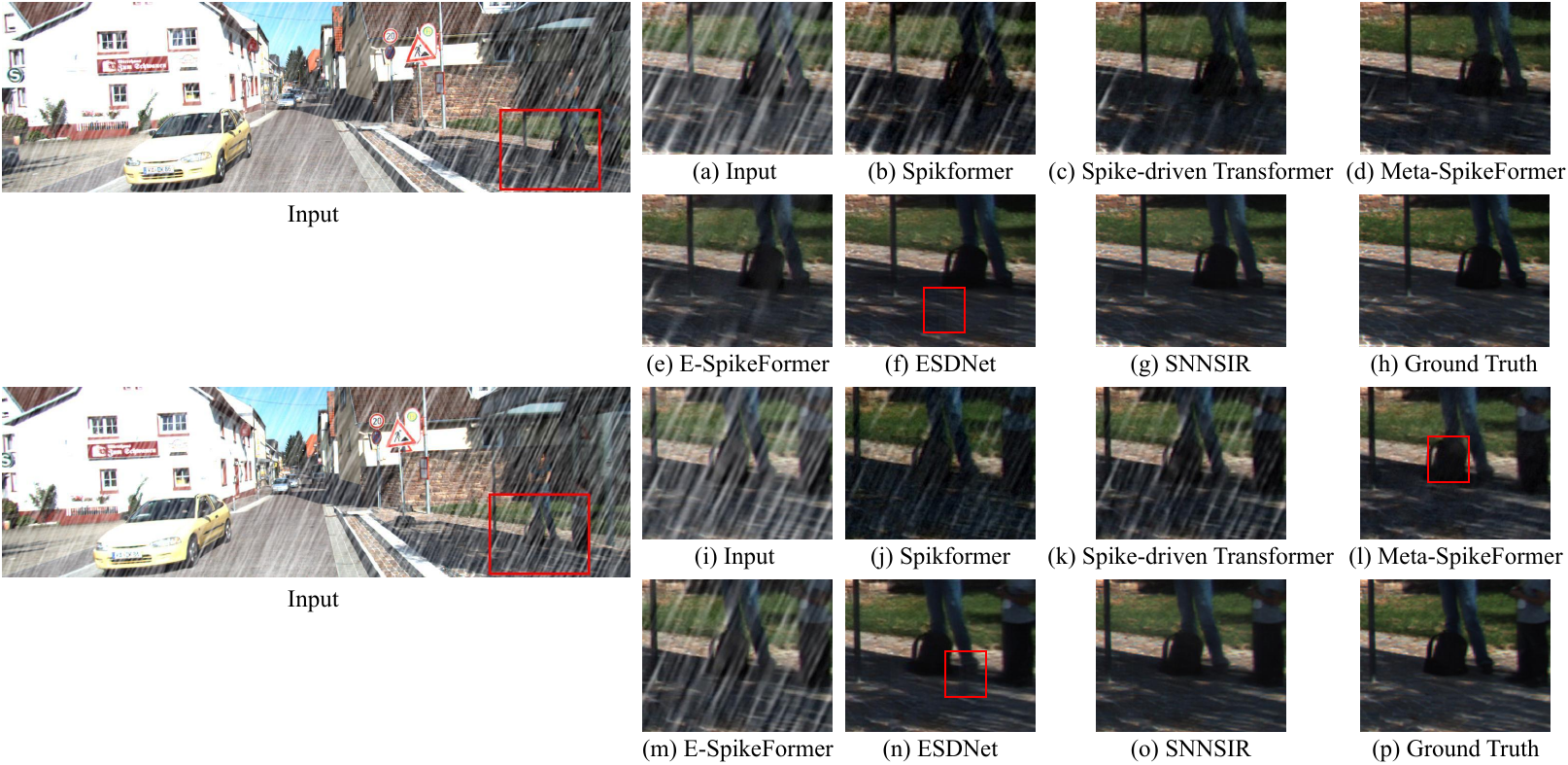}
	\caption{Visual comparison for rain streak removal on the RainKT15 dataset. The top and bottom rows show the left and right views, respectively. Best viewed by zooming in for details.}
	\label{fig:visualK15}
\end{figure*}

\begin{figure*}[t!]
	\centering
	\includegraphics[width=\linewidth]{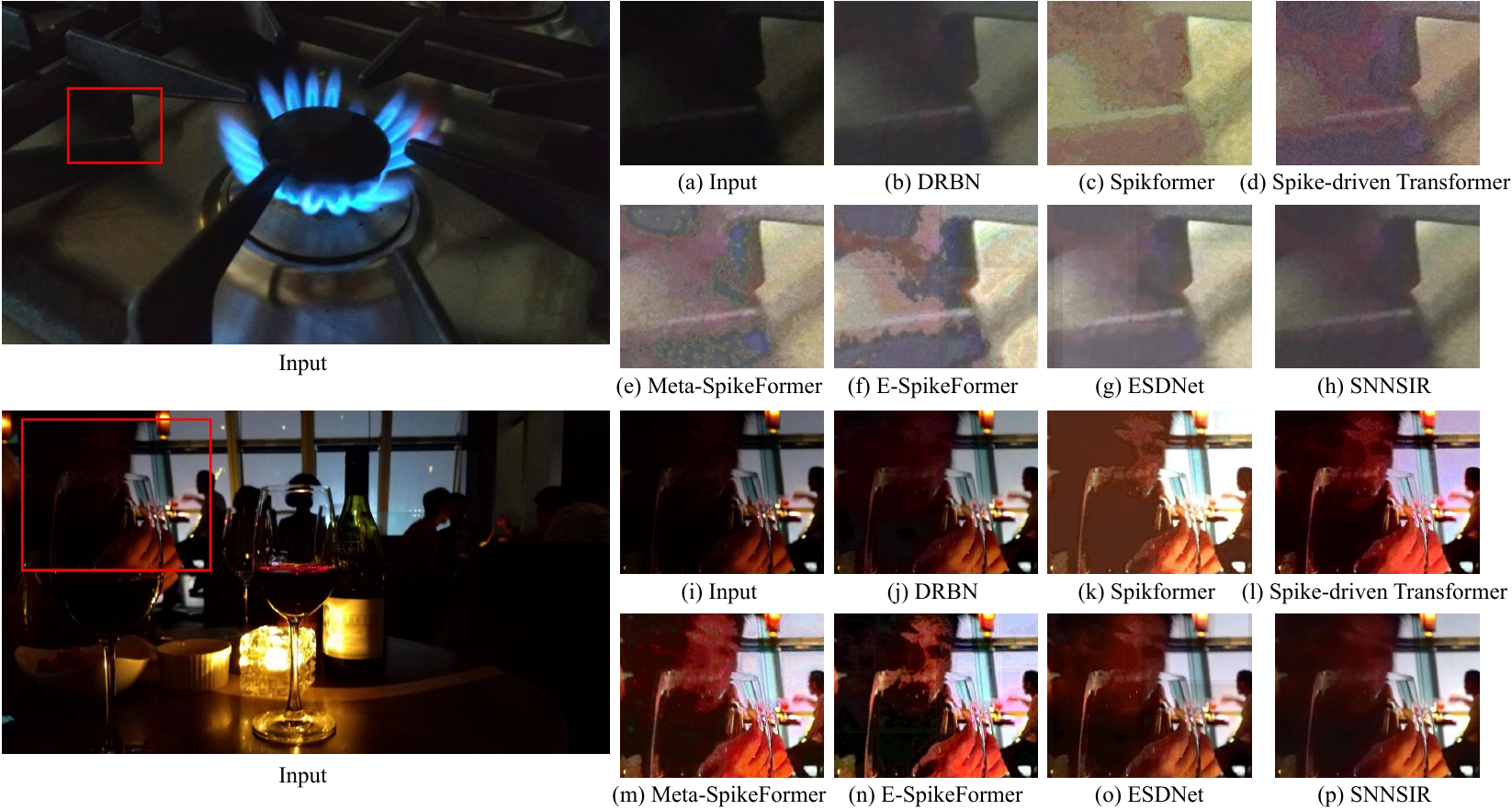}
	\caption{Visual comparison for low-light enhancement on the test real from the Holopix50k dataset. Best viewed by zooming in for details.}
	\label{fig:visualRealLowLight}
\end{figure*}


\begin{figure*}[t]
	\centering
	\includegraphics[width=\linewidth]{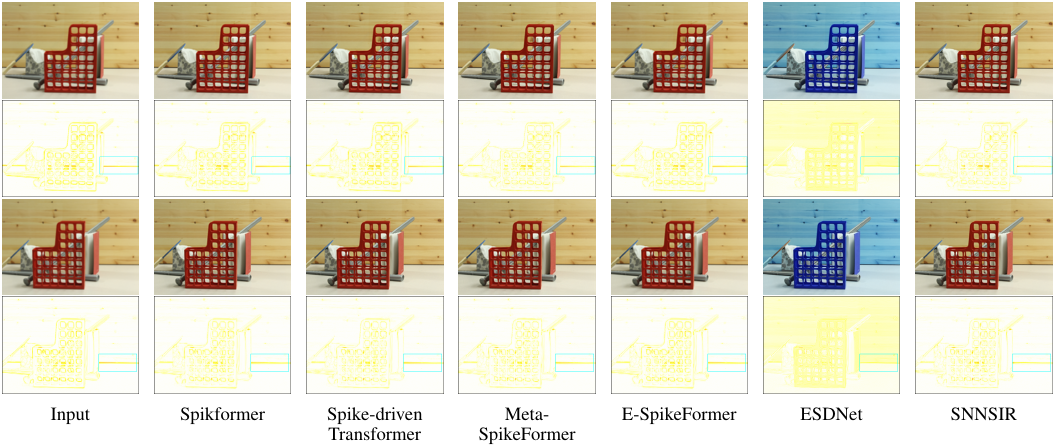}
	\caption{Visual comparison for super-resolution on the Middlebury dataset. The top and bottom rows show the left and right views, respectively. Best viewed by zooming in for details.}
	\label{fig:visualSR}
\end{figure*}

\begin{figure*}[t]
	\centering
	\includegraphics[width=\linewidth]{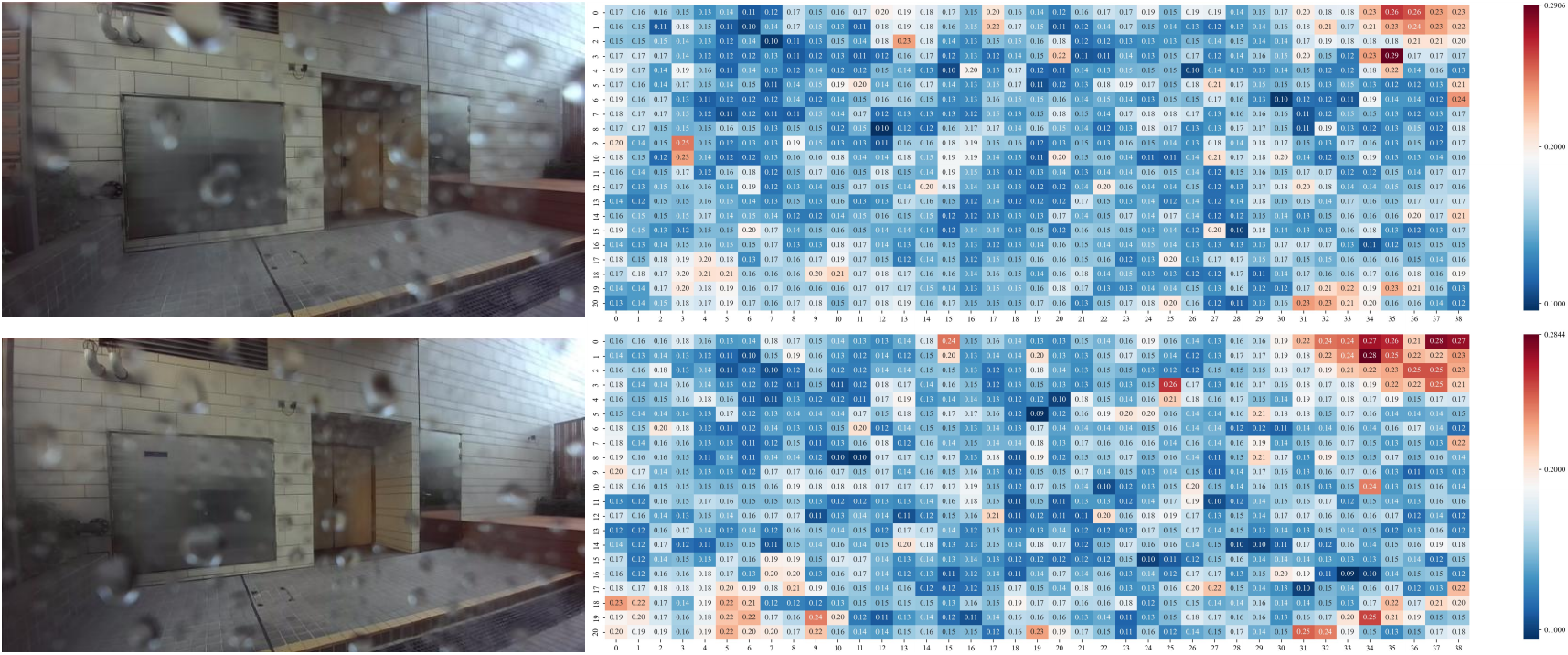}
	\caption{Spike firing rate maps of SNNSIR on the Stereo Waterdrop dataset. The top and bottom rows show the left and right views, respectively. The blue-white-red gradient represents a gradual increase in the firing rate.}
	\label{fig:firingMap}
\end{figure*}

\subsection{Stereo Image Deraining Results}
\label{sec:waterResults}
Table~\ref{tab:quantitiveStereo} presents the results of stereo image raindrop removal on the Stereo Waterdrop dataset~\cite{shi2021stereo}, while Table~\ref{tab:quantitiveKT} presents the results of rain streak removal on the RainKT12 and RainKT15 datasets~\cite{zhang2022beyond}.
Compared with ANN-based methods, SNNSIR outperforms the classical methods Restormer by 0.02 dB and RAttNet by 0.51 dB. Although its restoration performance differs negligibly from the former, it reduces the P by 87.52\% and E by 97.73\%. This demonstrates our proposed methods effectively leverage the spike-driven characteristics of SNNs, significantly reducing energy consumption, which shows
that our methods offer a fresh perspective for image restoration. Compared with SNN-based methods, SNNSIR significantly outperforms the other five baselines in both restoration performance and resource load across the two subtasks. Among these baselines, ESDNet, a monocular-specific restoration model, lacks advantages when handling binocular tasks. Furthermore, our proposed method addresses the absence of high-quality SNN-based baselines in binocular tasks via the reasonable use of non-linearity and cross-view interaction. Figure~\ref{fig:visualWaterdrop} shows that SNNSIR exhibits no prominent large-area red regions, with the overall color leaning more toward white, indicating superior performance in both detail and global restoration. In Figure~\ref{fig:visualK15}, among the results comparable to our proposed SNNSIR, we have highlighted the regions with poor restoration quality using red boxes. For ESDNet, noticeable vertical stripe artifacts appear in both the left and right views. As for Figure~\ref{fig:visualK15}(l), Meta-SpikeFormer fails to fully restore the black backpack region. The above issues do not occur in SNNSIR.

\subsection{Stereo Image Low-light Enhancement Results}
\label{sec:lowlightResults}
Experiments are conducted on the Middlebury~\cite{scharstein2014high} and Holopix50k~\cite{hua2020holopix50k} datasets. In Table~\ref{tab:quantitiveLLE}, SNNSIR outperforms SNN-based methods in terms of P, E and restoration quality. Specifically, it achieves superior PSNR and SSIM values with significantly lower energy consumption. Furthermore, even with a higher number of parameters, SNNSIR achieves performance comparable to DRBN while consuming only 20.99\% of its energy, demonstrating exceptional energy efficiency without compromising restoration quality. To evaluate the generalization ability, we measure the results on the real-world test set from the Holopix50k dataset. The enhancement results of the five SNN-based baselines exhibit overexposure or white artifacts, which are particularly noticeable on the foreheads of individuals in Figures~\ref{fig:visualRealLowLight}(k–o). In contrast, the results of SNNSIR appear more natural and visually consistent.

\begin{figure}[t!]
  \centering
  \includegraphics[width=\linewidth]{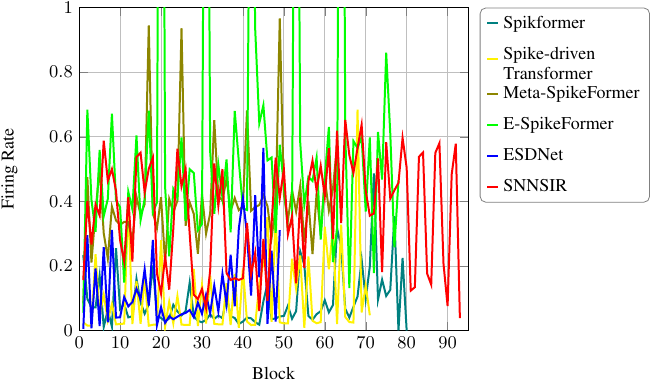}
  \caption{Spike firing rate of SNN-based baselines.}
  \label{fig:line_chart}
\end{figure}

\subsection{Stereo Image Super-Resolution Results}
\label{sec:SRResults}
We evaluate super-resolution~(SR) methods on the Middlebury\cite{scharstein2014high} and Flickr1024~\cite{wang2019learning} datasets. In line with the lightweight design philosophy of many SR methods, we adapt the five SNN baselines by removing up/downsampling and refinement operations, while maintaining a 32-channel width. The results in Table~\ref{tab:SRResults} indicate ESDNet struggles with this task. While other four baselines perform comparably, their quality is limited. Our SNNSIR surpasses them, achieving the highest restoration quality with significantly lower energy usage. In terms of visual comparison, SNNSIR yields results that are generally similar to those of several existing methods. However, the regions highlighted by blue boxes demonstrate that SNNSIR achieves superior detail preservation, as shown in Figure~\ref{fig:visualSR}. Notably, these baselines are not specifically designed for stereo SR, placing them at an inherent disadvantage. This reflects the current landscape of SNN-based stereo methods, establishing SNNSIR as the first tailored baseline for this task.

\subsection{Spike Firing Rate (SFR) Map}
We define the SFR as the average spike firing rate along the $T$-dimension of the spike neuron outputs in the network. Since each unit block in the network contains a single spike event, we use the block as the recording point. Through SFR, we can analyze the network's activation stability and spike distribution. Figure~\ref{fig:line_chart} illustrates that all SNN-based methods exhibit relatively low stability, yet they display a cyclical variation between high and low SFR, with SNNSIR showing this trend most prominently. This may be related to the activation and reset operations in SNNs, where high and low fluctuations in activation are expected. More appropriate SFR and more regular fluctuations contribute to better restoration quality and extremely low energy consumption. As shown in Figure~\ref{fig:firingMap}, the spike distribution in SNNSIR is mostly concentrated in the blue, low-activation regions, except for the top-right corner, which is influenced by highlights. This unexpected issue may suggest that SNNs are less suitable for low-light enhancement tasks, as low-light conditions hinder activation. The scattered distribution of high-activation regions (white and red) corresponds to the isolated raindrop points, indicating that SNNSIR can effectively focus on the raindrops.

\subsection{Ablation Studies} 
For convenience, all ablation experiments are carried out on the Stereo Waterdrop dataset, which contains a suitable number of samples and diverse scenes. 

\noindent\textbf{Impact of Different Designs.} Table~\ref{tab:albationCompnt} shows the effect of the key components. The Membrane Shortcut (MS)~\cite{hu2024advancing} used in SRBB brings better results to model (b) compared to model (a)~\cite{fang2021deep}. Similarly, a comparison between model (d) and model (e) demonstrates that SSRB yields an improvement of 0.68~dB in PSNR while introducing only 0.03~M additional parameters. Furthermore, compared to model (b), model (c) achieves a significant performance improvement, which demonstrates the effectiveness of the non-linearity introduced by SSCM after removing the activation function from the SNNs. We observe that model (f) results in a performance loss, indicating that activation functions involving exponential operations are contrary to the characteristics of spike-driven.

\noindent\textbf{Analysis of the Effect of SNN-Based Design.} To assess whether our SNN-based design alleviates the representational limitations of binary activations, we revert SNNSIR to an ANN-based version by replacing spike neurons with ReLU and adding activation functions to SSCM and SSCA. As shown in Table~\ref{tab:SNN2ANN}, the SNN-based model achieves both lower energy consumption and better performance. While image restoration is inherently a static task, the temporal evolution in SNNs facilitates iterative feature refinement and selective activation, contributing to more effective noise suppression and structural recovery.

\begin{table}[t]
	\caption{Ablation studies on the proposed components. $\sigma$ is the sigmoid function.}
	\label{tab:albationCompnt}
	\setlength{\tabcolsep}{1mm}
	\centering
	{
		\fontsize{9}{10}\selectfont
		\begin{tabular}{lccc}
			\toprule
			Model&P~(M)&PSNR &SSIM\\
			\midrule
			(a) SEW-RBB&\textbf{2.96}&23.03&0.824 \\
			(b) SRBB&\textbf{2.96}&24.96& 0.862\\
			(c) SRBB+SSCM&3.10&25.91&0.877 \\
			(d) SRBB+SSCA&3.09&25.10&0.862\\
			(e) SRBB+SSCA+SSRB&3.12&25.78&0.891 \\
			(f) SRBB+SSCM~($\sigma$)+SSCA+SSRB&3.26&\textbf{26.30}&\textbf{0.899}\\
			(g) SRBB+SSCM+SSCA+SSRB&3.26&\textbf{26.57}&\textbf{0.903}\\
			\bottomrule
		\end{tabular}
	}
\end{table}

\begin{table}[t]
	\caption{Comparison of SNN-based and ANN-based designs of the proposed SNNSIR.}
	\label{tab:SNN2ANN}
	\setlength{\tabcolsep}{1mm}
	\centering
	{
		\fontsize{9}{10}\selectfont
		\begin{tabular}{cccccc}
			\toprule
			Methods&P~(M)&PSNR&SSIM&E~(mJ)\\
			\midrule
			SNNSIR-ANN&3.26&26.48&0.900 &178.98\\
			SNNSIR-SNN&3.26&26.57&0.903 &29.32\\
			\bottomrule
		\end{tabular}
	}
\end{table}

\begin{table}[t!]
	\caption{The impacts of difference time steps.}
	\label{tab:differentTimeStep}
	\setlength{\tabcolsep}{1mm}
	\centering
	{
		\fontsize{9}{10}\selectfont
		\begin{tabular}{ccccc}
			\toprule
			Time steps & OPs~(G) &E~(mJ) & PSNR & SSIM \\
			\midrule
			1 & \textbf{3.008} & \textbf{4.38} & 21.84 & 0.821 \\
			2 & \textbf{14.795} & \textbf{15.83} & 25.94& 0.891 \\
			4 & 27.925 & 29.32  & \textbf{26.57}  & \textbf{0.903} \\
			8 & 56.660 & 58.54 & \textbf{26.42} & \textbf{0.901} \\
			\bottomrule
		\end{tabular}
	}
\end{table}

\noindent\textbf{Impact of Different Time Steps.} 
Table~\ref{tab:differentTimeStep} shows the effect of varying time steps on network performance. As the time step $T$ increases, performance initially improves, peaking at $T=4$, after which further increases lead to degradation. 
This reflects the inherent characteristic of neural networks, where increasing computation and parameters does not always yield gains, and aligns with the biological nature of SNNs, where excessive load may induce neural fatigue. Clearly, by comparing with the performance at $T=1$, it is evident that the performance improvement comes from the temporal dynamics of SNNs. However, even so, the theoretical increase in computational cost by a factor of T does not occur due to the spike-driven nature of SNNs.

\begin{table}[t!]
	\caption{FLOPs and SOPs of different SNN-based methods for different stereo image restoration tasks.}
	\label{tab:compution}
	\setlength{\tabcolsep}{1mm}
	\centering
	{
		\fontsize{9}{10}\selectfont
		\begin{tabular}{ccccc}
			\toprule
			Tasks& Methods & FLOPs~(G) & SOPs~(G) &E~(mJ)\\
			\midrule
			\multirow{9}{*}{\makecell[c]{Raindrop\\removal}}&Spikformer&0.566&416.498 &377.45\\
			& Spike-driven-&\multirow{2}{*}{0.566 }&\multirow{2}{*}{23.256}&\multirow{2}{*}{23.53}\\
			&Transformer& & & \\
			& Meta-&\multirow{2}{*}{0.566} &\multirow{2}{*}{283.480}&\multirow{2}{*}{257.74}\\
			& SpikeFormer& & & \\
			& E-SpikeFormer&0.566 &286.676&260.61\\
			& ESDNet&19.284 &95.469&174.63\\
			& SNNSIR&1.132 &26.793&29.32\\
			& SNNSIR-ANN&38.908 &0&178.98\\
			\midrule
			\multirow{8}{*}{\makecell[c]{Rain\\streak }}&Spikformer&0.566&1392.187&1255.57\\
			&Spike-driven- &\multirow{2}{*}{0.566}&\multirow{2}{*}{43.435}&\multirow{2}{*}{41.70}\\
			&Transformer & & & \\
			&Meta-&\multirow{2}{*}{0.566}&\multirow{2}{*}{220.940}&\multirow{2}{*}{201.45}\\
			& SpikeFormer& & & \\
			&E-SpikeFormer&0.566&310.342&281.91\\
			&ESDNet&19.284&142.697&217.13\\
			&SNNSIR&1.132&30.532&32.69\\
			\midrule
			\multirow{8}{*}{\makecell[c]{Low-light\\enhancement}}&Spikformer&0.566&537.319&486.19\\
			&Spike-driven- &\multirow{2}{*}{0.566}&\multirow{2}{*}{51.495}&\multirow{2}{*}{48.95}\\
			&Transformer & & &\\
			&Meta-&\multirow{2}{*}{0.566}&\multirow{2}{*}{331.878}&\multirow{2}{*}{301.30}\\
			& SpikeFormer& & & \\
			&E-SpikeFormer&0.566&498.600&451.34\\
			&ESDNet&19.284&133.155&208.55\\
			&SNNSIR&1.132&34.744&36.48\\
			\midrule
			\multirow{8}{*}{\makecell[c]{Super-\\resolution}}&Spikformer&0.046&8.956&8.27\\
			&Spike-driven- &\multirow{2}{*}{0.046}&\multirow{2}{*}{1.220}&\multirow{2}{*}{1.31}\\
			& Transformer& & & \\
			&Meta-&\multirow{2}{*}{0.046}&\multirow{2}{*}{19.031}&\multirow{2}{*}{17.34}\\
			& SpikeFormer& & & \\
			&E-SpikeFormer&0.046&56.175&50.77\\
			&ESDNet&0.385&21.277&20.92\\
			&SNNSIR&0.046&1.530&1.59\\
			\bottomrule
		\end{tabular}
	}
\end{table}

\section{Conclusion}
In this paper, we propose SNNSIR, a novel spiking neural network for stereo image restoration. Specifically, we address the absence of nonlinearity in SNNs through a structured modulation mechanism that simultaneously highlights degraded regions. To capture cross-view dependencies, we design a spike-compatible interaction strategy, while a lightweight refinement module is employed to recover fine-grained details. Experimental results on multiple stereo restoration tasks demonstrate that our model achieves competitive performance with significantly lower energy consumption, establishing a strong baseline for future research in SNN-based stereo image restoration.


 





\bibliographystyle{IEEEtran}
\bibliography{references}

\end{document}